\documentclass[letterpaper, 10 pt, conference]{ieeeconf}  

\IEEEoverridecommandlockouts                              

\usepackage{times}  
\usepackage{helvet}  
\usepackage{courier}  
\usepackage[hyphens]{url}  
\usepackage{graphicx} 
\usepackage{algorithm}
\usepackage{algorithmic}

\usepackage{adjustbox}
\usepackage{multirow}
\usepackage{subcaption}
\usepackage{caption}

\def\eg{\emph{e.g}., } 
 
\def\ie{\emph{i.e}., }

\def\etal{\emph{et al}.~}
\usepackage{newfloat}
\usepackage{listings}
\title{\LARGE \bf
Asset-Driven Sematic Reconstruction of Dynamic Scene with Multi-Human-Object Interactions
}

\author{Sandika Biswas$^{1,2}$ and Qianyi Wu$^{1}$ and Biplab Banerjee$^{2}$ and Hamid Rezatofighi$^{1}$ \\ 
$^{1}$ Monash University, Melbourne, Australia \\
$^{2}$ Indian Institute of Technology, Bombay, India \\
sandika.biswas@monash.edu}

\begin{document}

\maketitle

\begin{abstract}
Real-world human-built environments are highly dynamic, involving multiple humans and their complex interactions with surrounding objects. While 3D geometry modeling of such scenes is crucial for applications like AR/VR, gaming, and embodied AI, it remains underexplored due to challenges like diverse motion patterns and frequent occlusions. Beyond novel view rendering, 3D Gaussian Splatting (GS) has demonstrated remarkable progress in producing detailed, high-quality surface geometry with fast optimization of the underlying structure. However, very few GS-based methods address multihuman, multiobject scenarios, primarily due to the above-mentioned inherent challenges.
In a monocular setup, these challenges are further amplified, as maintaining structural consistency under severe occlusion becomes difficult when the scene is optimized solely based on GS-based rendering loss. 
To tackle the challenges of such a multihuman, multiobject dynamic scene, we propose a hybrid approach that effectively combines the advantages of 1) 3D generative models for generating high-fidelity meshes of the scene elements, 2) Semantic-aware deformation, \ie rigid transformation of the rigid objects and LBS-based deformation of the humans, and mapping of the deformed high-fidelity meshes in the dynamic scene, and 3) GS-based optimization of the individual elements for further refining their alignments in the scene. Such a hybrid approach helps maintain the object structures even under severe occlusion and can produce multiview and temporally consistent geometry. 
We choose HOI-M3 \cite{zhang2024hoi} for evaluation, as, to the best of our knowledge, this is the only dataset featuring multihuman, multiobject interactions in a dynamic scene. 
Our method outperforms the state-of-the-art method in producing better surface reconstruction of such scenes.
\end{abstract}


\section{Introduction}

Real-world environments built by humans are typically dynamic and involve multiple people in complex interactions with the elements of the surrounding scene. As a result, capturing the full spatial and semantic context through detailed 3D geometry recovery of entire scenes \cite{cao2023scenerf,schmied2023r3d3,han2025d,mao2024neural,kim2024omnisdf,wu2023objectsdf++,yu2022monosdf} has become a vital research direction, especially for applications such as gaming, embodied AI, robotics, and AR/VR—going beyond traditional object-centric surface reconstruction approaches \cite{yariv2021volume,wang2021neus,yariv2020multiview,oechsle2021unisurf,deng2022gram}. However, even though the Neural radiance field (NeRF) or Gaussian Splatting (GS)-based methods have shown excellent results in static scene reconstruction \cite{wu2023objectsdf++,yu2022monosdf,wu2022object}, very few studies focus on dynamic scenes with multiple humans and objects engaged in daily life interactions within a contextual environment. Few recent research \cite{biswas2024tfs,bhatnagar2022behave,huang2022intercap, Xie_2024_CVPR,xie2022chore} captures and models interactions of a single human with a single object in the scene. But, in reality, a real-world scene often involves complex interactions among multiple entities, \eg several humans engaging with various objects. Computer vision research has paid limited attention to this area, due to the complexities associated with capturing and representing such scenes.
\begin{figure*}[t]
    \centering
    \includegraphics[width=0.85\textwidth]{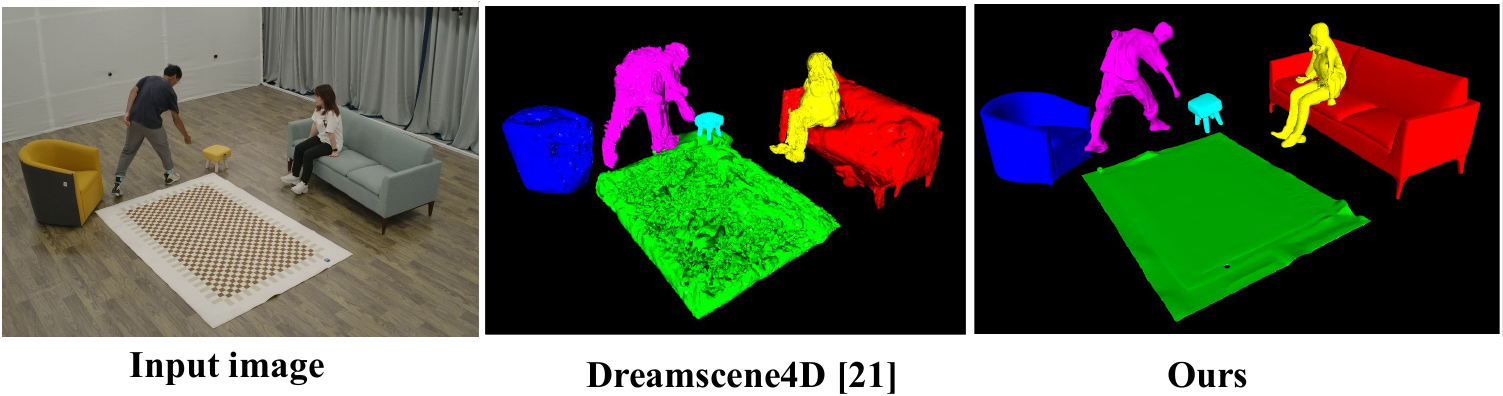}
    \caption{\small{\footnotesize{Semantic 3D geometry reconstruction of a dynamic scene with multiple human-object interactions (HOI-M3 \cite{zhang2024hoi} dataset). Our method can produce detailed, multiview-consistent, better-quality geometry with plausible overall scene reconstruction compared to the state-of-the-art method.} 
    }}
    \label{fig:teaser}
\end{figure*}

In this work, our objective is to produce semantic reconstruction while capturing the detailed 3D geometry of the scene elements from a single monocular video of such scenes where multiple entities are involved in daily life interactions, as illustrated in Fig. \ref{fig:teaser}. Few recent studies \cite{zheng2025gstar,dreamscene4d} with a similar setup either require dense multiview RGB-D videos to produce accurate, realistic reconstruction or fail to produce multiview consistent reconstruction while dealing with 
monocular RGB input. 
Close to our work, Dreamscene4D \cite{dreamscene4d}, which considers modeling a multi-object scene from monocular setup, proposes decomposing the scene into individual elements, and learning per element geometry by using Score Distillation Sampling (SDS) \cite{poole2022dreamfusion} to take advantage of a diffusion prior for generating multiview images of individual elements from the given input view. However, this method struggles to generate consistent multiview images, eventually affecting the quality of the reconstruction. Moreover, being solely dependent on per-frame GS-based optimization without leveraging the knowledge between the frames or the object semantics, it suffers from inaccurate geometry reconstruction under severe occlusion. 

To overcome these challenges, we propose a hybrid approach of 1) generating high-fidelity structures of individual elements of the scene, 2) deforming them or transforming them to get per-frame motions, and 3) finally fine-tuning their alignment in the scene using GS-based optimization, instead of purely GS-based optimization for learning their structure.  
Recently, many research studies have emerged in the field of 3D geometry generation \cite{xiang2024structured,tang2023dreamgaussian,liu2023one2345,liu2023one2345++,xu2024instantmesh} with the ability to generate high-quality, detailed 3D meshes from monocular images. These models leverage dense multiview visual features extracted from off-the-shelf, powerful vision foundation models that help generate consistent multiview content. We show that such a standard pre-trained 3D content generation model
can be used to create a strong initialization for the GS-based optimization. 
However, these models are not capable of generating scene-wise geometry and can produce high-fidelity structure only at the object level. Moreover, being reliant on a diffusion-based method, these models can not produce consistent geometry throughout the sequence.
Using the 3D generative models to produce a consistent geometry, with plausible interactions and alignments of the scene elements throughout the video, is not trivial. Modeling the motions of different elements is extremely challenging because of their diverse motions and severe occlusions. To mitigate these challenges, we design a pipeline where, given a monocular video, we first carefully choose frames (canonical frames) for every object in the scene, where the objects are fully visible, and predict 3D geometry with detailed surface by leveraging the pre-trained 3D generative models. To learn the motion of each object in the scene, we first decompose the scene into rigid and non-rigid scene elements. For non-rigid elements \eg humans, we use a pre-trained SMPL-based model \cite{patel2024camerahmr} to predict the pose for every frame and impose these per-frame poses on the canonical frame geometry. This helps us to correctly capture the motion of the human and get a very good initialization with detailed surface geometry for the entire sequence. For rigid elements, we align the generative model predicted meshes with a predicted point cloud of the scene \cite{wang2025vggt} for the canonical frame and impose per-frame motion by calculating the transformation between the point clouds of consecutive frames of the canonical frame. Finally, we initialize the Gaussians with the per-frame mesh vertices and optimize them using rendering loss to refine the 6DoF poses of the scene elements. 
Overall, our major contributions are as follows: 
\begin{itemize}
    \item We introduce ADSR, Asset-Driven Scene Reconstruction, that leverages 3D asset generation models for generating multiview consistent, high-quality, detailed surface/geometry reconstruction of scenes with multiple humans and multiple objects from monocular video.
    \item We propose a hybrid approach that combines the strengths of 3D generative models for generating high-fidelity scene elements, performs a coarse alignment of them in the scene, and finally, a GS-based optimization to refine their alignment.
    \item Our experiments show the benefits of this multi-modal fusion in terms of improved 4D prediction accuracy. Our model can reconstruct highly dynamic scenes and outperforms current state-of-the-art methods for multihuman, multiobject scene reconstruction.
\end{itemize}

\section{Related Works}
\textbf{3D reconstruction of human-scene interactions:} Several methods aim to model human-scene interactions and reconstruct human-object arrangements with varying degrees of generality and geometric details. POSA \cite{hassan2021populating}, PROX \cite{PROX:2019}, and CG-HOI \cite{diller2024cg} focus on predicting more plausible human poses with a given 3D scene, by leveraging the contextual knowledge of human-scene interactions, enabling more realistic predictions of how humans can exist and move within a given scene. PHOSA \cite{zhang2020phosa}, in contrast, emphasizes accurate spatial alignment between humans and objects given object meshes as input. Xie \etal  \cite{xie2022chore} and Weng \etal \cite{weng2021holistic} take a step further by jointly reconstructing humans and objects; however, they rely on template models rather than reconstructing the actual geometries and are limited in scope to a small set of object categories. A recent method \cite{wen2025reconstructing} first focuses on open vocabulary in-the-wild 3D HOI reconstruction, with the capability of generating any object reconstruction from a monocular image. However, these methods do not produce a detailed geometry of the human and focus on a static scene. In comparison, NeRF-based approaches like NeuralDome \cite{zhang2023neuraldome} and TFS-NeRF \cite{biswas2024tfs} enable high-fidelity geometry and novel view synthesis, but typically focus on a single human-object pair, limiting their applicability to more complex multi-object or full-scene scenarios.

\textbf{Dynamic scene reconstruction - dense pointmap generation:} With the rapid progress in foundational vision models, there have also been significant advancements in 3D scene reconstruction. One recent method, Dust3R \cite{wang2024dust3r}, leverages a Vision Transformer (ViT) backbone \cite{dosovitskiy2020vit} pre-trained with DINOv2 \cite{oquab2023dinov2} to extract highly informative image features, enabling robust generation of dense point clouds from multiview, unposed images of static scenes. Building on this, several works \cite{han2025d,zhang2024monst3r,chen2025easi3r,wang2025continuous,wang2025vggt} have extended the approach to handle scenes with dynamic motions, achieving high-quality dense pointmap generation even under complex temporal variations. 
For instance, Monst3R \cite{zhang2024monst3r} directly extends DUSt3R by training on dynamic scene datasets, while CUT3R \cite{wang2025continuous} introduces a recurrent updating mechanism within the transformer architecture to process video streams in an online, feed-forward manner. 
Despite these advancements, these methods generate pixel-aligned point maps that are inherently view-dependent, limiting their ability to reconstruct a complete 3D representation of the scene or underlying objects across arbitrary viewpoints. Comparably, we aim to generate a complete 3D surface reconstruction from monocular videos for such dynamic scenes. \\
\textbf{3D generative models:} With rapid progress in diffusion models, there has been a surge of research in 3D generative modeling. Early methods such as \cite{lin2023magic3d, poole2022dreamfusion,nichol2022point,jun2023shap} leverage text-to-image diffusion models to generate latent representations that guide the optimization of neural radiation fields (NeRFs), ultimately producing different forms of 3D output like textured meshes or point clouds. However, these approaches often suffer from slow optimizations and low-fidelity geometries due to the lack of multiview understanding of the given input. To address these limitations, more recent works generate multiview images using 2D diffusion models and use them to reconstruct 3D assets \cite{hong2023lrm,li2023instant3d,liu2024meshformer, liu2023zero,long2024wonder3d, shi2023mvdream, xiang20233d}. Despite improved efficiency, these 2D-assisted methods often produce lower-quality geometry due to inherent multiview inconsistencies in the generated images by 2D generative models. Some of the latest approaches \cite{liu2024one} aim to overcome these challenges by explicitly focusing on strategies for generating multiview consistent images. Liu \etal \cite{liu2024one} propose training a Stable Diffusion model to generate multiview consistent images, whereas Xiang \etal \cite{xiang2024structured} propose a Structured LATent (SLAT) representation that combines a sparse 3D grid with dense multiview visual features extracted from a powerful vision foundation model. These models can produce high-fidelity, multi-view consistent geometries. However, these models only work for a single image and may not produce consistent output for different frames of a video input, whereas we aim to utilize these models to produce a consistent geometry reconstruction for the entire sequence.

\begin{figure*}[ht!]
    \centering
    \includegraphics[width=0.95\textwidth]{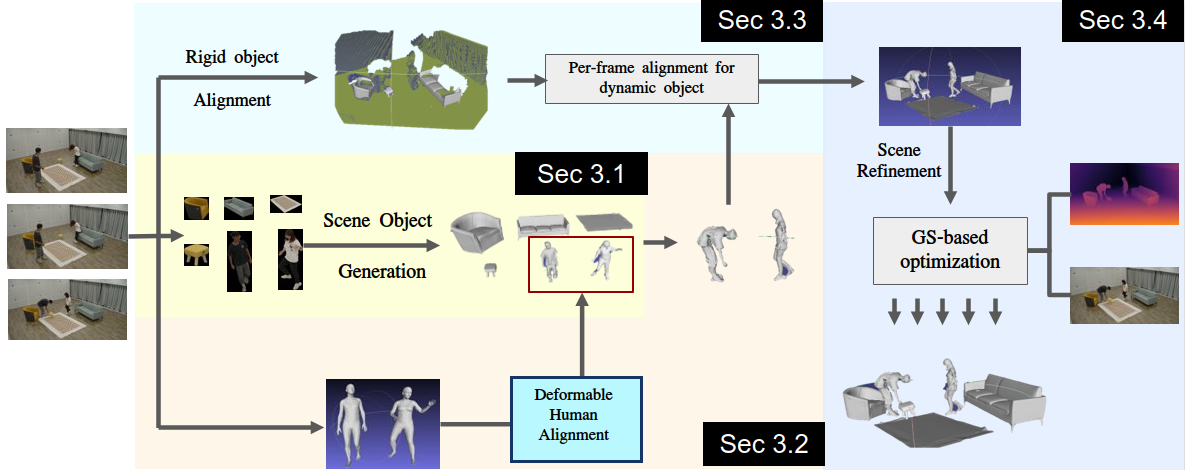}
    \caption{\small{\footnotesize{Overall pipeline of our method. Given a monocular video, we first generate the 3D mesh representation for every object (Sec 3.1), deform the human mesh (Sec 3.2), and transform the rigid objects (Sec 3.3) to align with every frame, and finally, fine-tune their alignment with GS-based optimization (Sec 3.4) to get a scene-level reconstruction.}}}
    \label{fig:bd}
\end{figure*}

\section{Methodology}
\textbf{Objective.} In the proposed method, we aim to produce high-fidelity geometry of a scene from a monocular video of multiple humans interacting with multiple objects. Fig. \ref{fig:bd} shows our overall pipeline with the following key components: 1) Generating the 3D mesh representation for every objects in the scene from chosen canonical frames (Sec. \ref{sec:trellis_init}), 2) Deforming the non-rigid object's mesh \eg humans, from the canonical frame to every other frame (Sec. \ref{sec:human_mapping}), 
3) Tracking and aligning the rigid object's mesh with every frame (Sec. \ref{sec:rigid_mapping}) through the sequence, 
4) Finally, optimizing these per-frame initializations of every object using Gaussian Splatting with rendering and depth loss for better alignment with the scene (Sec. \ref{sec:gs_opt}).

\subsection{High-fidelity 3D generation for scene objects:} 
\label{sec:trellis_init}
To produce high-fidelity geometry for the entire scene, it is necessary to first successfully capture the detailed geometry of the individual elements. This approach of decomposing the scene to the object level also helps disentangle the motions to better handle the complexity of the problem. DreamScene4D \cite{dreamscene4d} performs this object-level reconstruction while leveraging the multiview generative 2D diffusion models 
to first construct per-object meshes for a canonical frame, followed by using a deformation network to learn frame-specific deformations to reconstruct the motion. Although this is an effective approach to disentangle the object-wise motions, the overall quality of the reconstruction depends heavily on the fidelity of the canonical frame mesh. 
DreamScene4D employs Zero123 \cite{liu2023zero} to generate multiview images from a single reference, which often suffer from inconsistencies, leading to low-quality geometry reconstruction. Instead, we use a 3D generative model $\mathcal{G}$ \cite{xiang2024structured} that ensures multiview consistency by combining multiview 2D appearance features with 3D geometry. It defines a 3D grid for the underlying object and assigns dense multiview features from a powerful vision foundation model, enabling adequate multiview understanding that leads to producing high-quality 3D structure. 
However, this model fails to produce temporally consistent 3D geometries, hence, we utilize this model to generate meshes for every objects $\{\mathcal{O}_c\}^N_{i=1}$ in the scene $S$ only for respective canonical frames $\{I_{c}\}^N_{i=1}$ and carefully deform or transform them throughout the sequence. 
\begin{equation}
\mathcal{S} = \mathcal{G}(\{I_{c}\}^N_{i=1}) = \{\mathcal{O}_c\}^N_{i=1}
\end{equation}

\subsection{Temporally-coherent 3D Deformation of humans:} 
\label{sec:human_mapping}
\begin{figure} 
  \centering
  \includegraphics[width=0.45\textwidth]{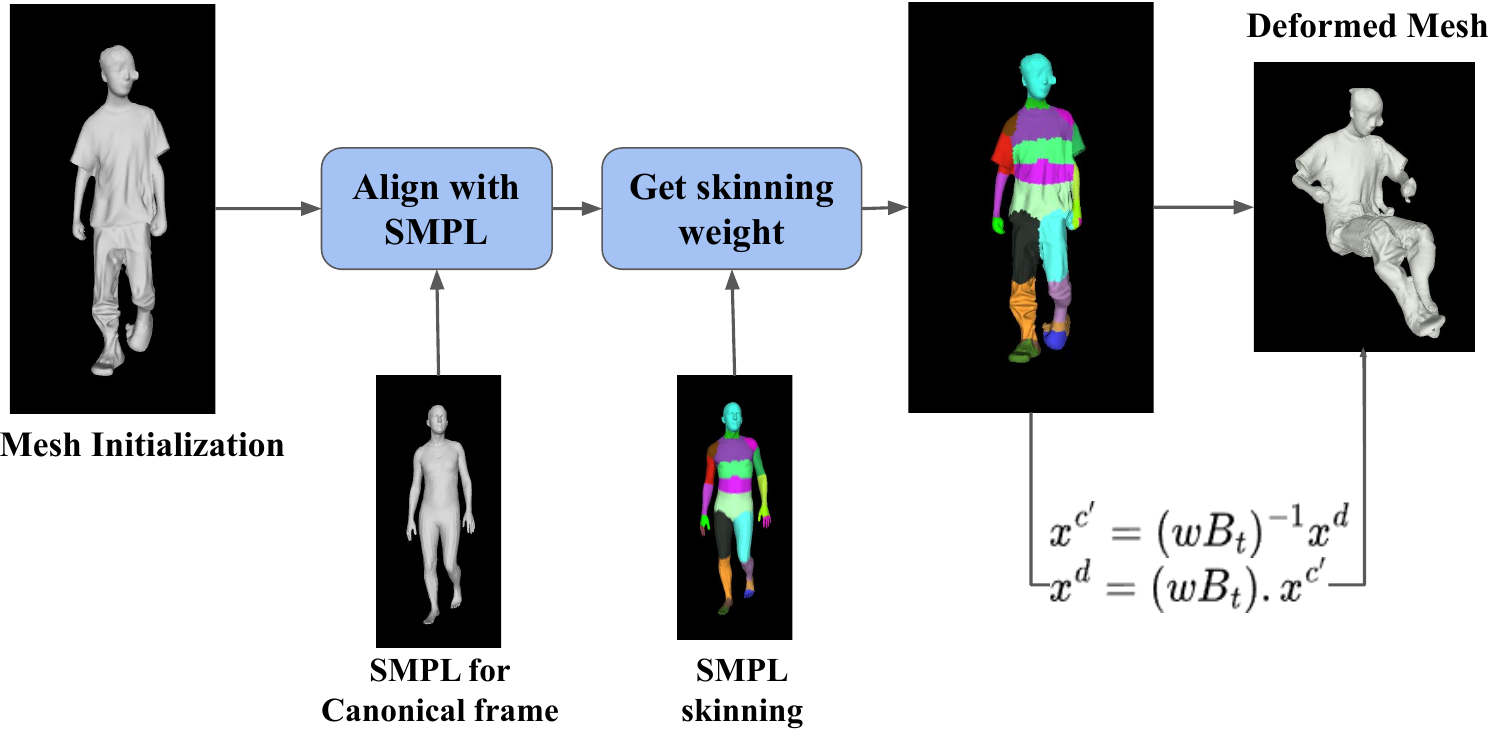}
  \caption{\small{\footnotesize{Deformation of the human mesh, generated by a 3D generative model, from the canonical frame to every other frame in the sequence.}}}
  \label{fig:smpl_deform}
  \vspace{-10pt}
\end{figure}

Learning the per-frame deformation from the canonical frame, solely based on GS-based optimization with per-frame rendering loss, makes it challenging to maintain the canonical frame geometry due to severe occlusions of the object instances under complex interactions. To tackle this challenge, we instead take advantage of foundational models to learn the deformations. For humans, we take the help of the Skinned Multi-Person Linear model (SMPL)\cite{loper2023smpl}, a widely used parametric 3D human body model that represents the human body shape and pose in a compact and parametric form. 
Building on SMPL, several NeRF and GS-based works \cite{peng2021animatable,weng_humannerf_2022_cvpr,guo2023vid2avatar,kocabas2024hugs} can reconstruct high-quality human meshes from sparse or monocular camera views, as SMPL gives a very good initialization for the optimization. Leveraging this fact, we aim to utilize the SMPL blend skinning weight ($\mathbf{w}$) and pose parameters ($\mathbf{\theta}$), for deforming the generative model predicted mesh $\mathcal{O}_c$ to produce temporally-coherent surface reconstruction for the dynamic humans throughout the sequence (Fig. \ref{fig:smpl_deform}). 
\vspace{-2pt}
The mapping between a SMPL T-pose mesh vertex ($x^T_c$)   (canonical pose for SMPL) and per-frame deformed mesh vertices ($x_d$) is defined as, 
\begin{equation}
x^d = \sum_{i=1}^{n_b} \mathbf{w} B_i x^c_T
\end{equation}
where $n_b$ represents the number of bones, $\mathbf{B}$ represents the bone transformation matrix derived from the pose parameters $\mathbf{\theta}$, and $\mathbf{w}$ represents the SMPL Linear Blend Skinning weight. Given that in our case, canonical pose ($x^c_v$) is chosen from the given video itself, rather than the SMPL defined T-pose ($x^c_T$), we first update the linear blend skinning representation to bridge the mapping between our chosen canonical frame and every other frame, by the following formulation, 
\begin{equation}
x^d_v = \mathbf{w}^{\prime}\mathbf{B^d_v} .(\mathbf{w}^{\prime}\mathbf{B^c_v})^{-1} x^c_v
\end{equation}
where $\mathbf{B^c_v}$ and $\mathbf{B^d_v}$ represent the bone transformation for the chosen canonical frame and target frame of the input sequence, respectively. $\mathbf{w}^{\prime}$ represents the skinning weight for the canonical mesh ($\mathcal{O}_c$) vertices $x^c_v$ generated in Sec. \ref{sec:human_mapping}. For each vertex in $x^c_v$, $\mathbf{w}^{\prime}$ is mapped from the SMPL skinning weight after aligning $x^c_v$ with the SMPL posed mesh for the canonical frame. Thus, the generative model predicted mesh $\mathcal{O}_c$ is deformed to every other frame.

\subsection{Tracking rigid objects through video frames:}
\vspace{-2pt}
\label{sec:rigid_mapping}
To map the predicted canonical meshes $\mathcal{O}_c$ (Sec. \ref{sec:rigid_mapping}) of the rigid objects within the scene, we first estimate per-frame depth using a monocular depth estimation model \cite{wang2025vggt}. By back-projecting the predicted depth maps for object-wise semantic masks, we obtain object-level point clouds $PC^{c,t}_{3D}$. First, the object-level point clouds for the canonical frames are used to scale and align the object-wise meshes $\mathcal{O}_c$ with the scene (using ICP registration).

Once the canonical meshes are aligned with the scene, we estimate the rigid transformations between two consecutive object-specific point clouds to transform the aligned canonical mesh to the next or previous frame. Specifically, we begin with the canonical frame point cloud $PC^{c}_t$ and compute its rigid transformations to the point clouds of adjacent frames: $PC^{c}_t \rightarrow PC_{t-1}$ and $PC^{c}_{t} \rightarrow PC_{t+1}$. These transformations are then applied to the scene-aligned meshes $\mathcal{O}_c$ to map them to the corresponding frames  $\mathcal{O}_{t-1}$ and $\mathcal{O}_{t+1}$ respectively. Similarly, $\mathcal{O}_{t-1}$ is transformed to $\mathcal{O}_{t-2}$ using the rigid transformation between $PC_{t-1} \rightarrow PC_{t-2}$. In this way, the dynamic rigid objects are tracked and mapped into the scene throughout the sequence.

\noindent
\subsection{Scene Refinement using Gaussian optimizer:} 
\label{sec:gs_opt}
The per-frame deformation of the scene generated through Section~\ref{sec:human_mapping}-\ref{sec:rigid_mapping} gives a good initialization of the reconstruction. However, the accuracy of the alignments of the individual scene elements in the scene depends on the accuracy of the predicted point cloud and the performance of the rigid registration, which may not produce perfect alignment with the input view. For this purpose, we choose further optimization of the initialized objects to refine their alignments in the scene. 
Recent advancements in neural rendering techniques, like 3D Gaussian splatting, have been proven to be very effective for learning 3D scene representations. Even though the primary focus for the GS-based techniques has been novel view synthesis, many recent studies have focused on improving the surface reconstruction of the underlying 3D geometry \cite{yu2024gsdf,turkulainen2024dnsplatter,lyu20243dgsr,wolf2024gsmesh}.
3DGS represents the scene as a set of 3D Gaussians ($G$) each represented by their mean ($\mu$), covariance matrix ($\sum$), opacity ($\alpha$) and color ($c$). 
\begin{equation}
G(p) = exp(-\frac{1}{2}(p-\mu)^T{\sum}^{-1}(p-\mu))
\end{equation}
where the covariance matrix $\sum$ is parameterized by scale ($S$) and rotation ($R$) of each Gaussians defined as $\sum = RSS^TR^T$.
These Gaussians are rasterized onto the image plane using a fast differentiable rasterizer, and finally, the color for the rendered image pixels is calculated as, 
\begin{equation}
C(x) = \sum_{i \in \mathcal{N}} c_i \alpha_i \prod_{j=1}^{i-1}(1 - \alpha_j)
\end{equation}
A rendering loss between the rasterized and the input image helps optimize the properties of the 3D gaussians \ie position ($\mu$), alignments ($R$), scale ($S$).

Although most GS-based 3D surface reconstruction methods focus on full-scene reconstruction, a few works \cite{cai2024gs,matteo20246dgs,wen2025reconstructing} have explored its application in 6D pose estimation of the objects. These studies have demonstrated that GS-based optimization can be more effective than traditional silhouette-based methods for aligning objects in a scene. 

Inspired by these methods, we utilize the Gaussian Splatting for refining the poses of the objects in the scene obtained through the above process explained in Sec. \ref{sec:trellis_init}-\ref{sec:rigid_mapping}. We initialize the Gaussian's mean at the vertices of $\{\mathcal{O}\}$ meshes.
Instead of optimizing Gaussian positions individually, we learn per-frame corrective transformations $\Delta R^t_i$ and $\Delta T^t_i$ for every object instance $i \in \{h,o\}$ in the scene. 
The objectwise Gaussians $G_{i}$, are updated as follows with the learnt corrective poses ($\Delta R, \Delta T$),
\begin{equation}
G^{\prime}_{i} = \Delta R^t_i G_{i} + \Delta T^t_i
\end{equation}

Gaussians are rasterized and supervised with rendering and depth losses to optimize the per-frame corrective transformations. Depth is rendered as,
\begin{equation}
D(x) = \sum_{i \in \mathcal{N}} z_i \alpha_i \prod_{j=1}^{i-1}(1 - \alpha_j)
\end{equation}
where $z_i$ is the z-coordinate of the Gaussians in the camera coordinate system. \\
\textbf{Training losses:}
The global orientation and translation of the object-wise Gaussians are optimized using the RGB rendering $\mathcal{L}_{color}$ and depth $\mathcal{L}_{depth}$ losses, 
\begin{equation}
    \mathcal{L}_{color} = (1 - \lambda_{ssim})||I_{gs} - I|| + \lambda_{ssim} · L_{ssim}(I_{gs}, I)
\end{equation}
where $I_{gs}$ and $I$ are rendered and input images, respectively. 
\begin{equation}
    \mathcal{L}_{depth} = ||\hat{D} - D||_2
\end{equation}
where $\hat{D}$ and $D$ are rendered and input depth images, respectively. The overall optimization function, 
\begin{equation}
    \mathcal{L} = \lambda_{color}\mathcal{L}_{color} + \lambda_{depth}\mathcal{L}_{depth}
    \label{equ:loss}
\end{equation}

\textbf{Implementation details:} The overall pipeline is implemented in Python using the PyTorch environment. For object-wise mesh initialization, we have used TRELLIS \cite{xiang2024structured}. 
For human motion prediction, we use the CameraHMR \cite{patel2024camerahmr} model, and evaluate it for every human track separately. 
To align the individual elements back into the scene, we take the help of a monocular depth prediction model, VGGT \cite{wang2025vggt}. 
In Gaussian Splatting, we follow the rasterization pipeline proposed in \cite{kerbl3Dgaussians} for rasterizing depth and RGB image for training with depth and rendering losses. We empirically set $\lambda_{D}$, and $\lambda_{R}$ to $10$ and $1$ respectively for Equ. \ref{equ:loss}.
\vspace{-10pt}

\section{Experiments}
\begin{table*}[t]
    \centering
    \begin{adjustbox}{max width=\textwidth}
    \begin{tabular}{l|llll|llll}
    \hline
        \multirow{3}{*}{\textbf{Methods}} & \multicolumn{4}{c}{\textbf{Obj-wise}} & \multicolumn{4}{c}{\textbf{Whole scene}} \\ \cline{2-9} 
         \multirow{2}{*}{} & \textbf{Dist.} $\downarrow$ & \textbf{Comp.} $\downarrow$ & \textbf{F-score} $\uparrow$ & \textbf{CD} $\downarrow$ & \textbf{Dist.} $\downarrow$ & \textbf{Comp.} $\downarrow$ & \textbf{F-score} $\uparrow$ & \textbf{CD} $\downarrow$ \\ 
         &  \textbf{Acc.} $(cm)$ & $(cm)$ &   $(\%)$ & $(cm)$ &  \textbf{Acc.} $(cm)$ & $(cm)$ & $(\%)$ & $(cm)$ \\\hline
        \textbf{Dreamscene4D} \cite{dreamscene4d} & 0.0734 & 0.0947 & 0.4801 & 0.0840 & 0.2165 & 0.2070 & 0.2272	& 0.2118 \\
        \textbf{Dreamscene4D} \cite{dreamscene4d} (trellis init) & 0.0679 & 0.0829 & 0.5095 & 0.0754 & 0.2187 & 0.2109 & 0.2237 & 0.2148 \\ 
        \textbf{Ours} & \textbf{0.0446} & \textbf{0.0539} & \textbf{0.6638} & \textbf{0.0493} & \textbf{0.0843} & \textbf{0.1185} & \textbf{0.3754} & \textbf{0.1014} \\ \hline
    \end{tabular}
    \end{adjustbox}
    \caption{\small{\footnotesize{Quantitative comparison of sobject-wise and whole-scene reconstruction quality on the HOIM3 dataset \cite{zhang2024hoi}. We compare the original Dreamscene4D \cite{dreamscene4d} with a variant where Gaussian means are initialized using TRELLIS-predicted mesh vertices (Dreamscene4D (Trellis init.)) instead of random initialization.}}}
    \label{tab:obj_hoim3}
\end{table*}
\begin{table*}[t]
    \centering
    \begin{adjustbox}{max width=\textwidth}
    \begin{tabular}{l|llll|llll} 
    \hline
         \multirow{2}{*}{\textbf{Methods}} & \multicolumn{4}{c|}{\textbf{Static scene}} & \multicolumn{4}{c}{\textbf{Dynamic scene}} \\ \cline{2-9} 
         \multirow{2}{*}{} & \textbf{Dist.} $\downarrow$ & \textbf{Comp.} $\downarrow$ & \textbf{F-score} $\uparrow$ & \textbf{CD} $\downarrow$ & \textbf{Dist.} $\downarrow$ & \textbf{Comp.} $\downarrow$ & \textbf{F-score} $\uparrow$ & \textbf{CD} $\downarrow$ \\ 
         &  \textbf{Acc.} $(cm)$ & $(cm)$ &   $(\%)$ & $(cm)$ &  \textbf{Acc.} $(cm)$ & $(cm)$ & $(\%)$ & $(cm)$ \\\hline
        \textbf{Dreamscene4D} \cite{dreamscene4d} & 0.0983	& 0.1384 & 0.3433	& 0.1183 & 0.0558	& 0.0670 & 0.5673 & 0.0613 \\  
        \textbf{Dreamscene4D} \cite{dreamscene4d} (trellis init) & 0.0965 & 0.1192 &	0.3547 & 0.1079  & 0.0506 & 0.0619 & 0.5999 &	0.0562 \\  
        \textbf{Ours} & \textbf{0.0464}	& \textbf{0.0623} & \textbf{0.5898}	& \textbf{0.0543} &  \textbf{0.0422}	& \textbf{0.0487} & \textbf{0.7183}	& \textbf{0.0454} \\ \hline 
    \end{tabular}
    \end{adjustbox}
    \caption{\footnotesize{Quantitative comparison of static vs dynamic scene reconstruction quality on HOI-M3 \cite{zhang2024hoi} dataset. Here, the dynamic scene contains both human and dynamic rigid objects.}}
    \label{tab:obj_static_dyn_hoim3} 
\end{table*}

\textbf{Baseline:}
We use DreamScene4D \cite{dreamscene4d} as the primary baseline for comparison in our multi-human, multi-object dynamic scene setup. DreamScene4D tackles novel view synthesis for dynamic scenes involving multiple objects by decomposing the scene into individual elements and reconstructing each separately using a GS-based optimization framework. It models dynamics by factorizing motion into global camera motion and independent object motions. After learning the object-specific motions, it remaps the reconstructed objects into the full scene using a depth-based loss. For our experiments, we utilize their publicly available implementation to optimize the scenes in our setup. The original method includes an optional inpainting step to complete the object masks in regions where occlusions occur. However, due to the complexity of our scenes, characterized by multiple humans and objects with prolonged and severe occlusions, it is not feasible to obtain such inpainted images for every instance. Therefore, we evaluate DreamScene4D without inpainting.

To ensure a fair comparison, we also introduce a modified baseline with two key changes to the original \cite{dreamscene4d} setup: 1)  In the original pipeline, DreamScene4D learns the geometry of each object by optimizing the Gaussian representation using the first frame of the sequence. In our case, objects may be occluded in the first frame. Hence, we instead supply a canonical frame, where the object is fully visible for learning the canonical structure. 2) DreamScene4D initializes each object's geometry from a randomly sampled point cloud. In contrast, we enhance this initialization by leveraging meshes predicted by TRELLIS, allowing for a more informed starting point for Gaussian optimization.

\noindent
\textbf{Dataset and evaluation metrics:} To the best of our knowledge, HOI-M3 \cite{zhang2024hoi} is the only dataset to capture multi-human multi-object interactions in a contextual setup, where both humans and objects are in motion. We choose 5 sequences from the dataset, \textit{livinigroom$\_$data01}, \textit{livingroom$\_$data12}, \textit{livingroom$\_$data36}, \textit{office$\_$data02}, \textit{office$\_$data15}, for our experimental setup. These sequences have 2-3 humans interacting with 4-5 objects. 
For every sequence, we choose 600-700 frames from the whole sequence, where there are substantial interactions or movements between the objects.

Following the previous approaches \cite{wu2022object,wu2023objectsdf++}, we report distance accuracy (\textit{Dist. Acc.}), completeness (\textit{Comp.}), F-score, and Chamfer distance (\textit{CD}) for evaluating the quality of the geometry reconstruction (Tab. \ref{tab:obj_hoim3}). We have taken a similar approach as \cite{biswas2024tfs} to evaluate both scene element-wise reconstruction quality as well as the whole scene reconstruction quality. 
Also, we give an analysis of the reconstruction quality of static scene (static objects) vs dynamic scene (dynamic objects + dynamic humans) (Tab. \ref{tab:obj_static_dyn_hoim3}). For the whole scene (Tab. \ref{tab:obj_hoim3}), we first align the predicted scene with the ground-truth scene by performing an ICP registration using only the static elements of the scene. This analysis quantifies the quality of the overall alignment of the scene elements.
\begin{figure*}[t]
    \centering
    \includegraphics[width=0.9\textwidth]{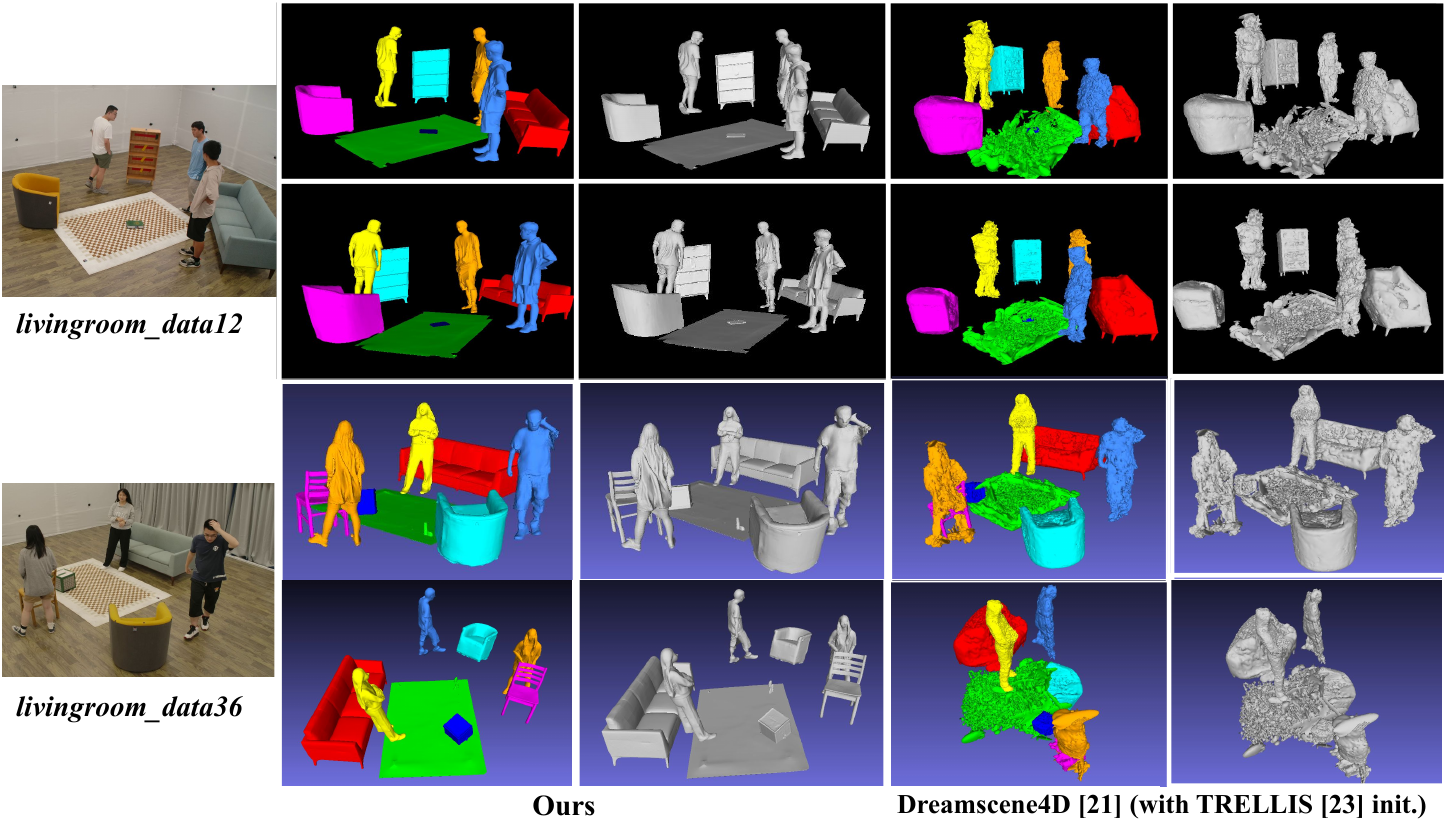}
    \caption{\footnotesize{Reconstruction results for two scenes,  \textit{livingroom$\_$data12} and \textit{livingroom$\_$data36} of the HOI-M3 dataset \cite{zhang2024hoi} with multi-human multi-object interactions. For every example, we show the reconstruction results from two different views. Our method can produce detailed, multiview-consistent, better-quality geometry with plausible overall scene reconstruction.} 
    }
    \label{fig:qual}
\end{figure*}

\noindent
\textbf{Results:} In this section, we present a comprehensive analysis of the performance of our method in geometry reconstruction in comparison with the SOTA. In Tab. \ref{tab:obj_hoim3}, we present quantitative results for individual object reconstruction as well as whole scene reconstruction averaged over all the sequences. 
Our method outperforms the SOTA for all metrics.

\begin{figure}[t!]
    \centering
    \includegraphics[width=0.49\textwidth]{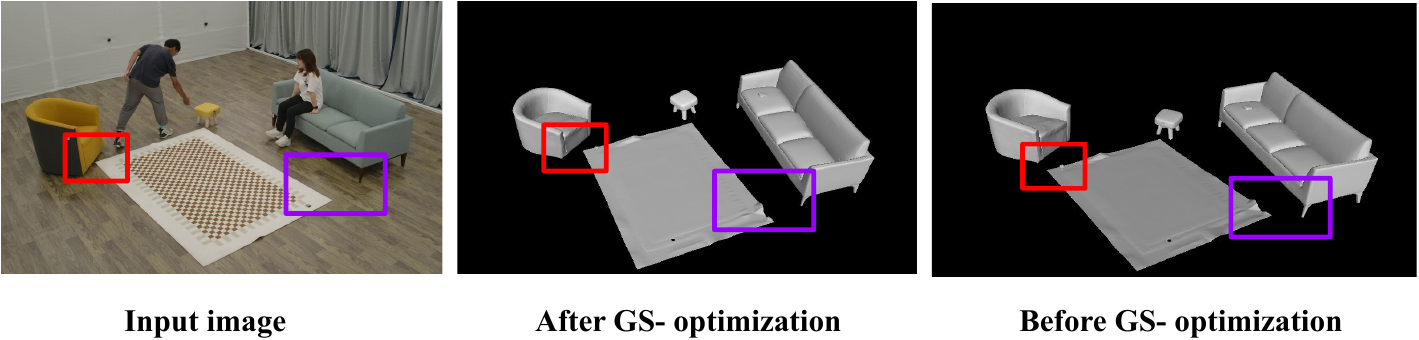}
    \caption{\footnotesize{(a) Input image, (b), (c) An instance of object alignments before and after Gaussian Splatting-based optimization. Areas under the marked rectangles show misalignment between the scene elements before GS-based optimization, which improves with GS-based optimization.}}
    \label{fig:abl_gs_vs_trellis}
    \vspace{-10pt}
\end{figure}

It is worth noting that when Dreamscene4D was evaluated using TRELLIS mesh initialization (2nd row, Tab. \ref{tab:obj_hoim3}) instead of randomly initialized Gaussians (1st row, Tab. \ref{tab:obj_hoim3}), the reconstruction quality improved. However, the results were still hindered by their per-frame Gaussian optimization with view-inconsistent multiview images. It is evident from Fig. \ref{fig:qual}, which shows that Dreamscene4D can generate reasonable geometry when visualized from the input view, while resulting in implausible geometries from the other views.
In contrast, our method achieves better reconstruction by explicitly leveraging the semantic information of the scene elements to guide the motion learning, instead of optimizing the Gaussians for every frame to learn the per-frame structure. Specifically, we estimate the transformation between consecutive frames from the predicted point clouds (Sec. \ref{sec:rigid_mapping}) and apply this transformation to mesh initialization $\mathcal{O}_c$ of the rigid objects. For human subjects, we use Linear Blend Skinning (LBS) to deform them from their canonical space to each frame. This approach preserves the structural integrity of both rigid and non-rigid elements, maintaining the temporal and multiview consistency throughout the sequence. This leads to our better performance for both static and dynamic scene reconstruction (Tab. \ref{tab:obj_static_dyn_hoim3}) compared to the SOTA.\\
\noindent
\textbf{Ablative study:} In this section, we provide an ablation for different components of our overall pipeline.\\
\textit{\textbf{A. Importance of GS-based optimization:}} As we can see in Tab. \ref{tab:ablative study}, even though our mesh initialization (Sec \ref{sec:trellis_init} - \ref{sec:rigid_mapping})
can achieve high-quality scene reconstruction, our 3DGS-based optimization further improves the performance with better alignment of the scene elements. Fig.\ref{fig:abl_gs_vs_trellis} shows an example of static scene reconstruction before and after GS-based optimization. GS-based optimization improves the alignments of the scene elements. \\
\noindent
\textit{\textbf{B. Object alignment vs surface reconstruction:}}
For element-wise accuracy evaluation, we adopt two approaches: 1) \textit{With ICP registration:} We align each predicted mesh to its corresponding ground-truth mesh using ICP. This removes global misalignment and allows us to assess the quality of geometry reconstruction independently of scene placement. 2) \textit{Without ICP:} We directly compare the predicted and ground-truth meshes without ICP alignment. This evaluates both the reconstruction accuracy and the correctness of the object’s orientations in the scene, thereby capturing the overall element-wise alignment quality (Tab. \ref{tab:ablative study}). It shows that even though our scene alignment performs better than that of the state-of-the-art method, it still has scope for improvement.\\
\noindent
\textit{\textbf{C. Sensitivity to Depth Prediction Accuracy:}}
We further analyze how the accuracy of the underlying depth or point cloud prediction methods influences the performance of our reconstruction model. In particular, even when these models yield relatively inaccurate or sparse reconstructions (Cut3r \cite{wen2025reconstructing} produces inaccurate depth compared to VGGT \cite{wang2025vggt}), our approach maintains consistent performance in capturing the surface geometry, as our model does not rely on per-pixel depth but instead relies on the median depth of the object.
\begin{table}[!ht]
    \centering
    \begin{adjustbox}{max width=0.48\textwidth}
    \begin{tabular}{l|llll} 
    \hline
         \multirow{2}{*}{\textbf{Methods}} & \multicolumn{4}{c}{\textbf{Whole scene}} \\ \cline{2-5} 
         \multirow{2}{*}{} & \textbf{Dist. Acc.} $\downarrow$ & \textbf{Comp.} $\downarrow$ & \textbf{F-score} $\uparrow$ & \textbf{CD} $\downarrow$ \\ 
         & $(cm)$ & $(cm)$ & $(\%)$ & $(cm)$  \\\hline
        \textbf{Ours w/o GS} & 0.0382 & 0.0409	& 0.7356 & 0.0396 \\  
        \textbf{Ours w GS} & \textbf{0.0378} & \textbf{0.0366} & \textbf{0.7509} & \textbf{0.0372} \\\hline
        \textbf{Dreamscene4D w/o ICP reg.} & 0.1068	& 0.1146& 0.3437 & 0.1107\\  
        \textbf{Dreamscene4D w ICP reg.} & 0.0826 & 0.1044 & 0.4412 & 0.0935\\

        \textbf{Ours w/o ICP reg.} & 0.0497 & 0.0506	& 0.6322 & 0.0501 \\   
        \textbf{Ours w ICP reg.} & \textbf{0.0382} & \textbf{0.0409} & \textbf{0.7731} & \textbf{0.0396} \\\hline    
        \textbf{Ours w VGGT + w/o ICP reg.} & 0.0497 & \textbf{0.0506}	& \textbf{0.6322} & \textbf{0.0501} \\  
        \textbf{Ours w Cut3r + w/o ICP reg.} & \textbf{0.0492} & 0.0535 &	0.6242 &	0.0514
 \\ \hline 
    \end{tabular}
    \end{adjustbox}
    \vspace{-3pt}
    \caption{\small{\footnotesize{Ablation study on \textit{livingroom$\_$data01} of HOIM3 \cite{zhang2024hoi} dataset.}}}
    \label{tab:ablative study}    
\end{table}
\\
Extended ablations and video results are included in the supplementary material.

\section{Conclusions}
In this work, we introduce a hybrid approach for the semantic reconstruction of complex dynamic scenes from monocular RGB videos with multihuman, multiobject interactions.
We propose decomposing the problem into the following key components: generating high-fidelity meshes for individual objects, using foundational models for learning deformation and transformations of non-rigid and rigid objects, aligning the individual instances with the scene, and fine-tuning the overall scene alignment using GS-based optimization. In this approach, our method can produce temporally-coherent, multiview consistent high-fidelity geometry for every object in the scene throughout the video, with a better alignment of the scene objects compared to the existing approach.

\textbf{Limitations and future work:} Our approach relies on foundational models for human motion prediction, which may introduce pose inaccuracies, addressable via GS-based optimization. Additionally, collision handling between objects remains unexplored, and we plan to investigate this further.


\bibliographystyle{IEEEtran}
\bibliography{aaai2026}

\end{document}